%% file: ELALR-Completeness.tex
 \documentclass{amsart}
 \usepackage[active]{srcltx} 

\usepackage{times}

\usepackage{caption}
\usepackage{tabularx}
\usepackage{setspace}

 \usepackage{latexsym}   
\usepackage{subfig}
\usepackage{graphicx}

\usepackage{listings}

 \newsavebox{\savepar}

  \title{Minimizing LR(1) State Machines is NP-Hard}
  \author{Wuu Yang \\
          National Yang-Ming Chiao-Tung University, Taiwan, R.O.C.}
  \date{\today. This work is supported, in part, by Ministry of Science and Technology, Taiwan,
R.O.C., under contracts MOST 108-2221-E-009 -050 -MY3.
A preliminary version of this paper appeared in Proc. 14th International Conf. Autonomic and Autonomous Systems (ICAS 2018) May 20-24, 2018, Nice, France}

\begin{document}


\title{\textbf{\Large Minimizing LR(1) State Machines is NP-Hard}}



\begin{abstract}
 LR(1) parsing was a focus of extensive research in the past 50 years.
 Though most fundamental mysteries have been resolved, a few remain hidden in the dark corners.  
 The one we bumped into is the minimization of the LR(1) state machines, which we prove is NP-hard. 
 It is the node-coloring problem that is reduced to the minimization puzzle. 
 The reduction makes use of two technique: indirect reduction and incremental construction.
 Indirect reduction means the graph to be colored is not reduced to an LR(1) state machine {\it directly}.
 Instead, it is reduced to a context-free grammar from which an LR(1) state machine is derived.
 Furthermore, by considering the nodes in the graph to be colored one at a time,
 the context-free grammar is incrementally extended from a template context-free grammar 
 that is for a two-node graph.
 The extension is done by adding new grammar symbols and rules.
 A minimized LR(1) machine can be used to recover a minimum coloring of the original graph.
     \footnote{
     This work is supported, in part, by Ministry of Science and Technology, Taiwan, R.O.C., 
     under contracts MOST 103-2221-E-009-105-MY3 and MOST 105-2221-E-009-078-MY3.
     }
 
 
\end{abstract}

     \maketitle

\section{Introduction}

 Parsing is a basic step in every compiler and interpreter.
 LR parsers are powerful enough to handle almost all practical programming languages \cite{Knuth1965}.
 The downside of LR parsers is the huge table size.
 This caused the development of several variants, such as LALR parsers, 
 which require significantly smaller tables at the expense of reduced capability.

           \begin{figure}
           \begin{center}
            \resizebox{4.0in}{!}{\includegraphics{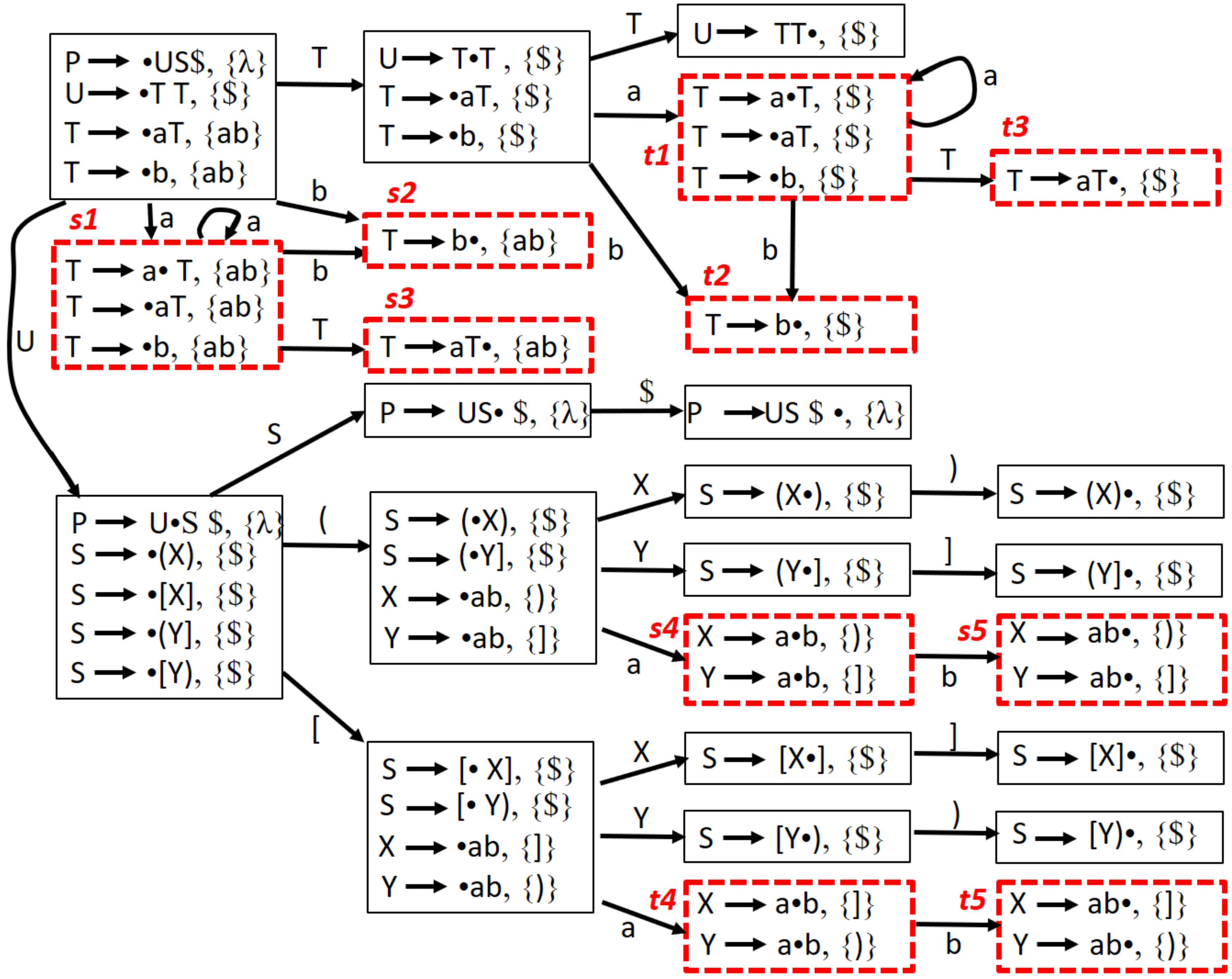}}
            \caption{The LR(1) machine of a grammar. \label{G3}}
            \end{center}
           \end{figure}

 The core of an LR(1) parser is a deterministic finite state machine.
 The LALR(1) state machine may be obtained by merging {\it every} pair of similar states\footnote{Two
     states are {\it similar} if and only if they become identical if the look-ahead sets in the items
     in the two states are ignored.} 
 in the LR(1) machine \cite{Fischer2009}.
 In case (reduce-reduce) conflicts occur due to merging,\footnote{Only reduce-reduce conflicts
          may occur due to merging similar states.}
 the parser is forced to revert to the larger, original LR(1) machine.
 A practical advantage of LALR(1) grammars is the much smaller state machines
 than the original LR(1) machines.
 However, any conflicts will force the parser to use the larger LR(1) machine.

 If every pair of similar states in an LR(1) state machine are merged together (ignoring conflicts),
 the resulting LR(1) state machine would be isomorphic to the LR(0) machine \cite{Fischer2009}\cite{Aho2006}.
 Due to the significant size difference between the two state machines, we know there are many
 pairs of similar states in an LR(1) machine.
 If any pair of similar states may cause conflicts, the parser will be forced to use 
 the much larger LR(1) machine.
 It would be more reasonable to merge some, but not all, pairs of similar states \cite{Yang2017}.
 The result, called an {\it extended LALR(1) state machine}, would be a state machine that is smaller 
 than the LR(1) machine but larger than the LALR(1) machine.

\begin{figure}
           \begin{center}
            \resizebox{4.0in}{!}{\includegraphics{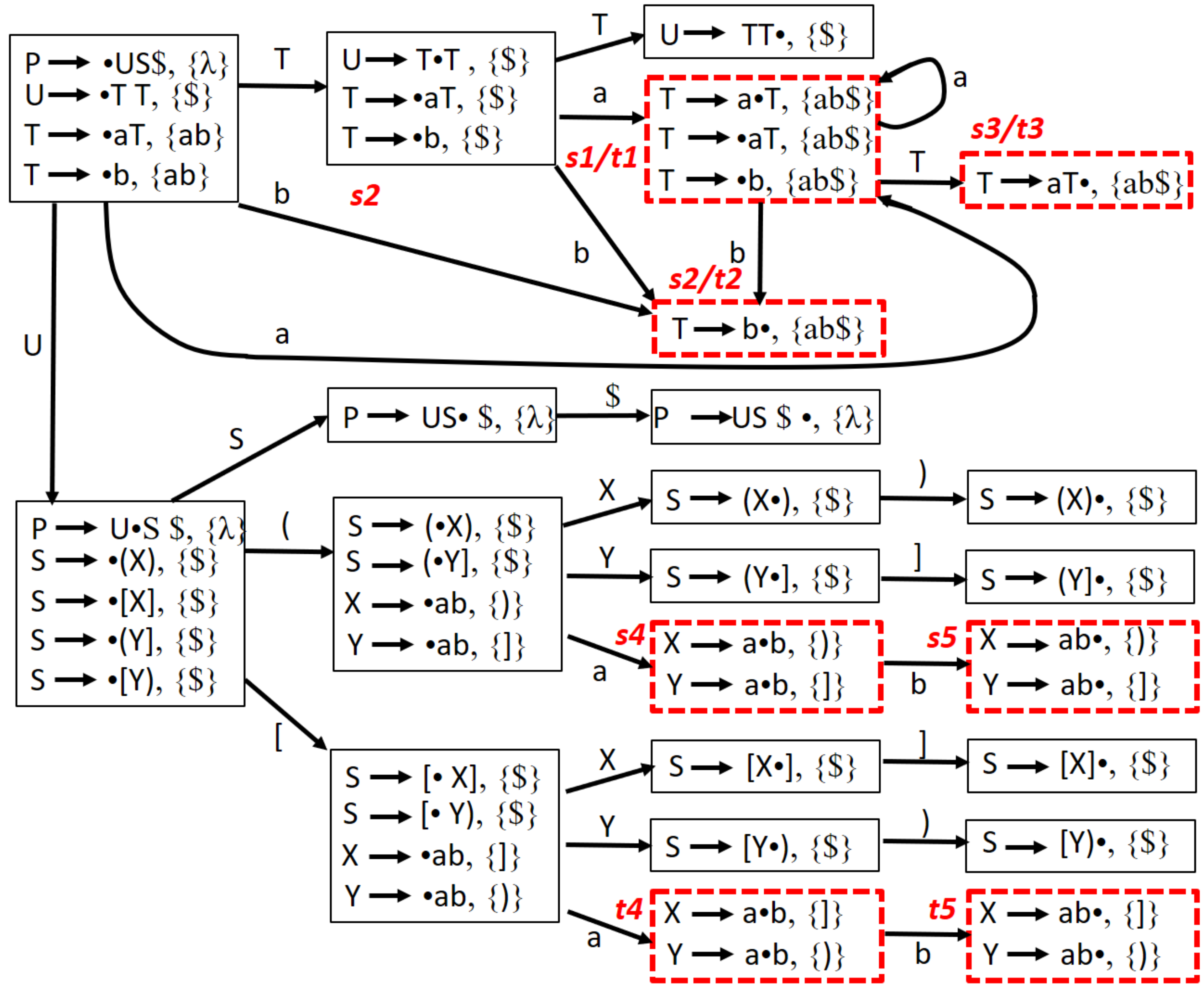}}
            \caption{The corresponding minimum LR(1) machine for Figure \ref{G3}. \label{G3-reduced}}
            \end{center}
           \end{figure}

 For example, there are five pairs of similar states in the LR(1) machine in Figure \ref{G3}.
 Only three pairs---$(s_{1}, t_{1}), (s_{2}, t_{2}), (s_{3}, t_{3})$---can be merged.
 The pair of similar states---$(s_{5}, t_{5})$---cannot be merged due to a (reduce-reduce) conflict.
 The last pair of similar states---$(s_{4}, t_{4})$---cannot be merged because $(s_{5}, t_{5})$ are not merged 
 for otherwise the resulting machine would become nondeterministic.
 Figure \ref{G3-reduced} is the corresponding (minimum) LR(1) machine.

 In general, two states in an LR(1) machine can be merged if and only if the following two conditions
 are satisfied:
	\begin{enumerate}
	\item The two states must be similar;
	\item Corresponding successor states of the two states must have already been merged.
	\end{enumerate}

 A further question is if there is an efficient algorithm that can merge the {\it most}
 number of similar states, thus producing a {\it minimum} LR(1) state machine.
 That is, we wish to minimize the LR(1) state machine.
 Since the number of similar states is finite, a na\"ive approach is to try all possibilities.

 Our study shows that minimizing the LR(1) state machine is an NP-hard problem.
 We reduce the node-coloring problem to this minimization problem.
 Starting from an (undirected) graph to be node-colored, we construct a context-free grammar.
 Then the LR(1) machine of the context-free grammar is derived.
 We can use an algorithm to calculate the corresponding minimum LR(1) machine.

 In order to recover a minimum coloring from this minimum LR(1) machine, we can perform one more easy step.
 In the LR(1) machine, every state that is not similar to any other states is removed, 
 leaving only similar states.
 Then an edge between two similar states is added if the two similar states may cause conflicts.
 The resulting machine is called a {\it conflict graph}.
 Merging similar states in the LR(1) machine is essentially identical to merging states in the 
 conflict graph.\footnote{Due to the construction of the grammar,
                      all states in the resulting graph are similar to one another in the LR(1) machine.
                      Furthermore, the conflict graph is actually isomorphic to the original color graph.}
 From the minimum LR(1) machine, it is straightforward to recover a minimum coloring.

 The following theorem seems obvious but we wish to bring it to the reader's attention when reading this paper:

 {\bf Theorem}.
 {\it Let $s_{1}$, $s_{2}$, and $s_{3}$ be three states in an LR(1) machine.
 If the three states are not conflicting pairwise, then merging all three states
 will not create any conflicts.}

 Due to the above theorem, we need to consider only pairs, not triples, quadruples, {\it etc}., of similar states.
 This greatly simplifies our discussion.

 Note also that there might be more than one minimum LR(1) machine for a given LR(1) machine.


 LR parsers were first introduced by Knuth \cite{Knuth1965}.
 Since LR parsers are considered the most powerful and efficient practical parsers, 
 much effort has been devoted to related
 research and implementation \cite{Aho1974},\cite{Anderson1973},\cite{DeRemer1971},\cite{Johnson1978},\cite{Mickunas1976},\cite{Pager1977}.

 The canonical LR(1) parsers make use of big state machines.
 For some LR(1) state machines, {\it every} pair of similar states may be merged.
 These are called LALR(1) state machines (or LALR(1) parsers) \cite{DeRemer1969}. 
 One way to generate LALR(1) machines is to merge all pairs of similar states.
 On the other hand, since an LALR(1) machine is isomorphic to the corresponding LR(0) machine,
 many algorithms are proposed to add the look-ahead information to the states in the LR(0) machine
 in order to obtain the LALR(1) machine \cite{Anderson1973},\cite{DeRemer1982},\cite{Pager1977}.
 None of these attempted to parse LR(1)-but-non-LALR(1) grammars. 

 It is known that every language that admits an LR($k$) parser also admits 
 an LALR(1) parser \cite{Mickunas1976}.
 In order to parse for an LR(1)-but-non-LALR(1) grammar, there used to be four approaches:
 (1) use the much larger LR(1) parser;
 (2) add ad hoc rules to the LALR(1) parser to resolve conflicts, similar to what yacc \cite{Johnson1978} does;
 (3) merge some, but not all, pairs of similar states \cite{Yang2017}; and
 (4) transform the grammar into LALR(1) and then generate a parser.
 The transformation approach may exponentially increase the number of production rules \cite{Mickunas1976}
 and the transformed grammar is usually difficult to understand.
 This paper shows that, although we wish to merge as many pairs of similar states as possible, this optimization problem is
 NP-hard.

 Pager proposed two methods: ``practical general method'' (PGM) \cite{Pager1977} and ``lane-tracing method''  
 \cite{Pager1973} \cite{Pager1977Lane}.
 The PGM method conceptually starts from the complete LR(1) machine and attempts to merge similar states.
 In constructing the LR(1) machine, when a new state is generated, PGM attempts to merge the new state with an
 existing state if a (strong or weak) {\it compatibility test} is passed.
 The compatibility test essentially determines if further derivation of the LR(1) machine will cause conflicts.
 On the other hand, the lane-tracing method starts from LR(0) machine. 
 When a conflict is detected, the relevant states are split in order to eliminate the conflicts.
 Splits continue until all conflicts are resolved.
 Chen \cite{Chen2011} actually implements Pager's two methods as well as other improvements, 
 such as unit-production elimination.
 Even though we proved the minimization problem is NP-hard, it is still important to build practical parser generators
 in the LR family.
 Practically, Pager and Chen's work is one of the best existing LR parser generators.

 The IELR method \cite{Denny2010} includes additional capability to eliminate conflicts even if the grammar is not LR.
 For instance, IELR can handle certain ambiguous grammars, similar to {\tt yacc}r.

 Both \cite{Pager1977} and \cite{Denny2010} attempt to find a {\it minimal} machine.
 However, {\it minimal} simply means ``very small'' or ``locally minimum''
 rather than ``globally minimum''\cite{Denny2010}.
 This is different from our study of {\it minimization}.
 



 The remainder of this paper is organized as follows.
 Section 2 will introduce the terminology and background.
 Section 3 introduces a reduction algorithm that translates an undirected graph into a context-free grammar
 and discusses the reduction of the coloring problem to the minimization problem.
 The last section concludes this paper.

           \begin{figure}
           \begin{center}
            \resizebox{4.5in}{!}{\includegraphics{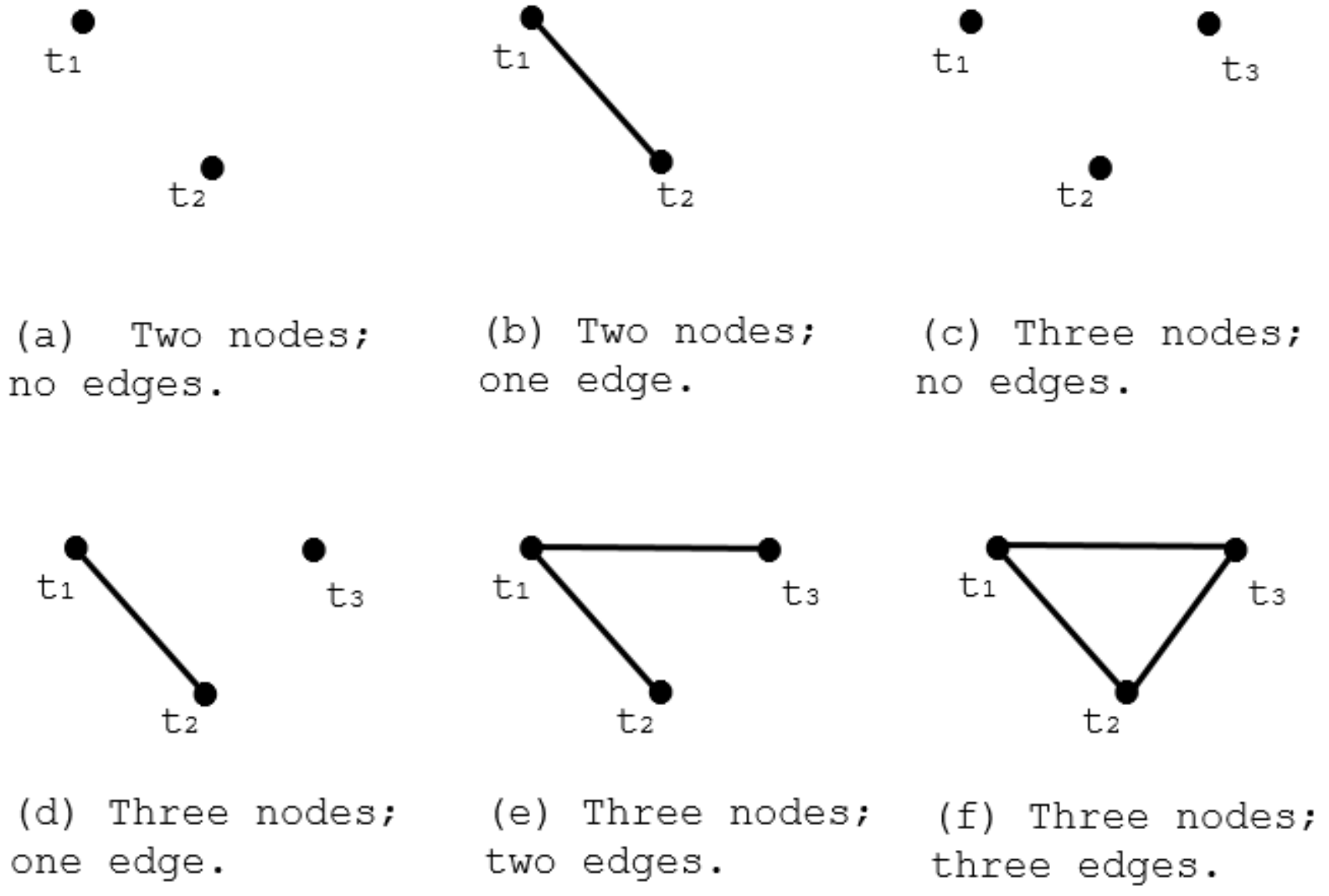}}
            \caption{Two cases for a color graph with two nodes and  
             four cases for a color graph with three nodes.\label{G33}}  
            \end{center}
           \end{figure}

 \section{Terminology and Background}

 A grammar $G = (N, T, P, S)$ consists of a non-empty set of nonterminals $N$, a non-empty
 set of terminals $T$, a non-empty set of production rules $P$ and 
 a special nonterminal $S$, which is called the start symbol.
 We assume that $N \cap T = \emptyset$.
 A production rule has the form 
	\begin{center}
  A ::= $\gamma$
	\end{center}
 where $A$ is a nonterminal and $\gamma$ is a (possibly empty) string of 
 nonterminals and terminals.
 We use the production rules to derive a string of terminals from the start symbol.

           \begin{figure}
           \begin{center}
            \resizebox{4.5in}{!}{\includegraphics{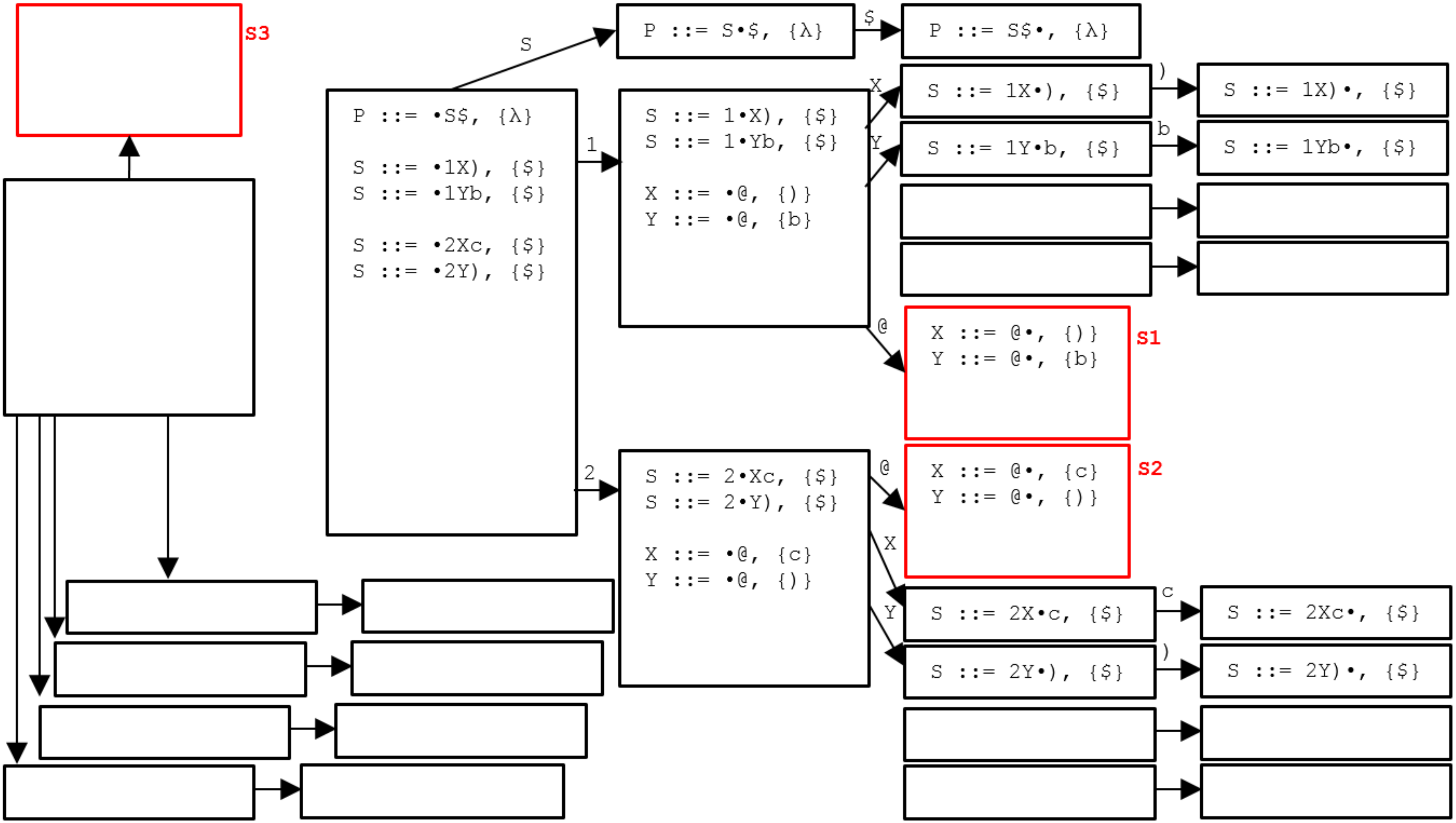}}
            \caption{The LR(1) machine for the graph in Figure \ref{G33} (b).
                     Note that there is a single conflict $s_{1} \leftrightarrow s_{2}$. 
                     The empty boxes are not part of this machine.
                     They are used for comparison with later machines.\label{G32-1}}  
            \end{center}
           \end{figure}

 LR(1) parsing is based on a deterministic finite state machine, called the LR(1) machine.
 A state in the LR(1) machine is a non-empty set of items.
 An {\it item} has the form (A ::= $\gamma_{1}$ $\bullet$ $\gamma_{2}$, $la$),
 where A ::= $\gamma_{1} \gamma_{2}$ is one of the production rules, 
 $\bullet$ indicates a position in the string $\gamma_{1} \gamma_{2}$, and 
 $la$ (the {\it lookahead} set) is a set of terminals that could 
 follow the nonterminal A in later derivation steps.
 The algorithm for constructing the LR(1) state machine for a grammar is explained in most compiler
 textbooks, for example, \cite{Aho2006}\cite{Fischer2009}.
 An example state machine is shown in Figure \ref{G3}.

 Two states in the LR(1) machine are {\it similar} if they have the same number of items and
 the corresponding items differ only in the lookahead sets.
 For example, states $s1$ and $t1$ in Figure \ref{G3}, each of which contains three items,
 are similar states.  
 Similarity among states is an equivalence relation.
 On the other hand, conflicting between two states is not transitive.

           \begin{figure}
           \begin{center}
            \resizebox{4.5in}{!}{\includegraphics{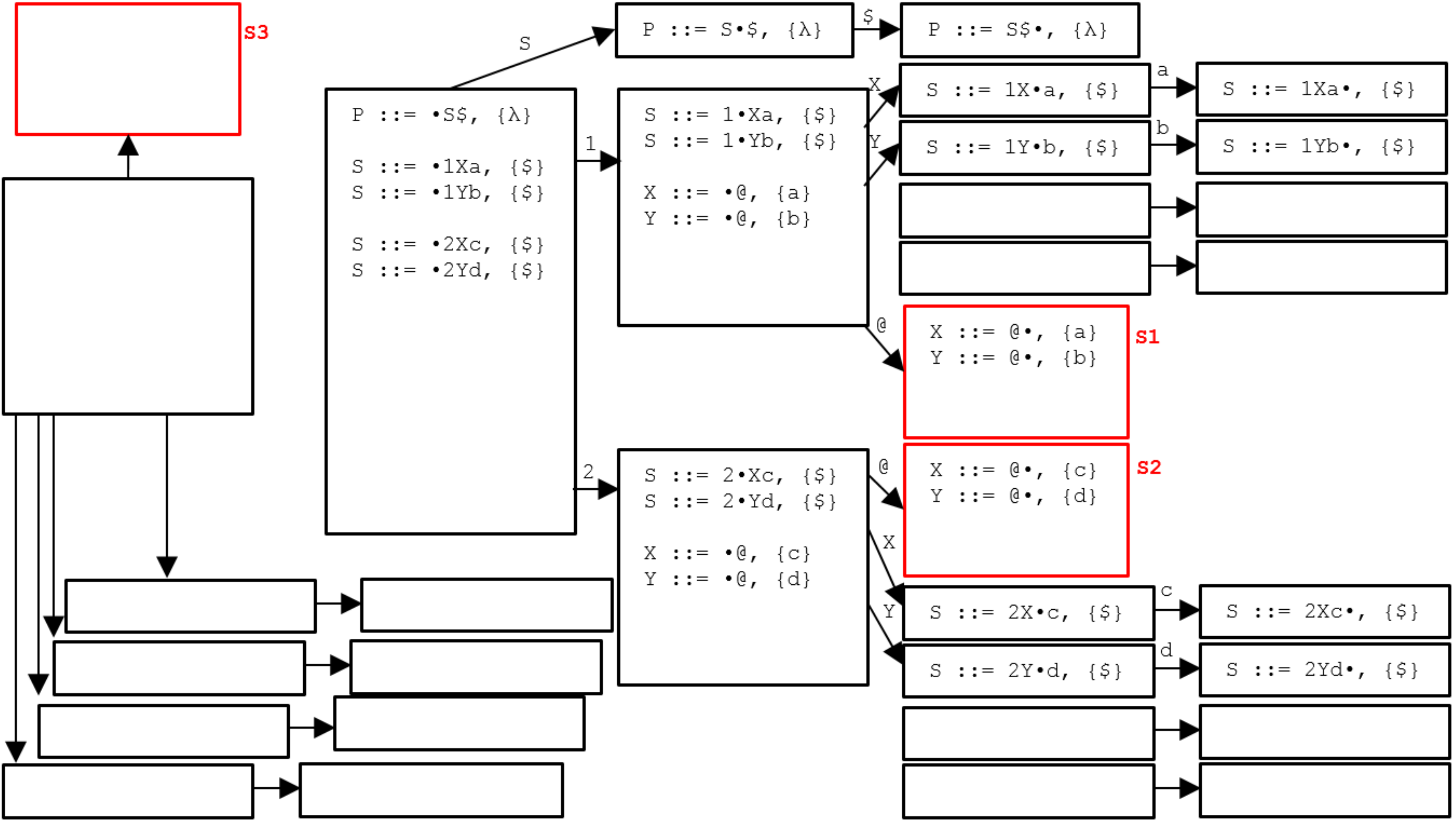}}
            \caption{The LR(1) machine for the graph in Figure \ref{G33} (a).
                     There is no conflict in the machine.
                     The empty boxes are not part of this machine.
                     They are used for comparison with later machines.  \label{G32-0}}  
            \end{center}
           \end{figure}

 LR(1) state machines are closely related to LR(0) state machines.
 However, an LR(1) machine is much larger than the corresponding LR(0) machine because many
 similar states are introduced.
 In order to reduce the size of the LR(1) state machine, some or all pairs of similar states may be {\it merged}\footnote{
    Remember two states are similar if they have the same items, except that the lookahead sets might differ.
    To {\it merge two similar states}, we use the same items in the original states,
    except that the lookahead set of an item is the union of the lookahead sets of the two
    corresponding items in the original states.
    }
 as long as no conflicts occur. 
 For example, LALR(1) machines are obtained from LR(1) machines by merging {\it every} pair of similar states.

           \begin{table}
           \begin{center}
           \caption{The grammar on the left corresponds to the color graph in Figure \ref{G33} (a).
           The grammar on the right corresponds to the color graph in Figure \ref{G33} (b).\label{table2}}   
           \begin{tabular}{|l|l|} \hline
 (no edge)  &   (one edge) \\
 P ::= S\$  &  P ::= S\$  \\
   &    \\
 S ::= 1Xa  &  S ::= 1X)  \\
 S ::= 1Yb  &  S ::= 1Yb  \\
   &    \\
 S ::= 2Xc  &  S ::= 2Xc  \\
 S ::= 2Yd  &  S ::= 2Y)  \\ 
   &    \\
 X ::= @    &  X ::= @     \\  
 Y ::= @    &  Y ::= @     \\   \hline
           \end{tabular}
           \end{center}
           \end{table}

 Sometimes merging two similar states may create a {\it (parsing) conflict}.
 For example, when states $s1$ and $s2$ in Figure \ref{G32-1} are merged, the resulting state
 will contain two items: (X ::= @ $\bullet$, $\{$ ), c $\}$) and (Y ::= @ $\bullet$, $\{$ b, ) $\}$).
 Because the terminal ``)'' appears in two different items in which $\bullet$ appears 
 at the end of the right-hand side, this is a parsing conflict.

 The aim of {\it minimizing an LR(1) machine} is to merge as many pairs of similar states as possible without
 causing conflicts.
 Our study shows that this minimization problem is NP-hard.

 \section{Reduction}

 We may prove that minimizing LR(1) machines is an NP-hard problem
 by reducing the node-coloring problem to this minimization problem.
 Specifically, from a graph $F$ to be colored, we construct a context-free grammar $G$.
 Then the LR(1) state machine $M$ is derived from $G$.
 An algorithm is used to calculate the minimum state machine, from which a minimum coloring can be recovered.

 In order to recover a minimum coloring, $M$ can be simplified by
 removing every state that is not similar to any other state,
 resulting in a {\it conflict graph}.
 Merging similar states in the conflict graph is essentially identical to finding a minimum coloring of $F$.

 We define a {\it node-coloring of a graph} as a partition of the set of nodes in the color graph satisfying 
 the requirement that
 nodes connected by an edge cannot be in the same partition block.
 A {\it minimum coloring} is a partition with the fewest blocks.
 Similarly, a {\it merge scheme} of an LR(1) state machine is a partition of the states satisfying the requirement that
 states in the same partition block are similar to one another and do not conflict with one another.
 A {\it minimal merge scheme} is a partition with the fewest blocks.
 {\it Minimizing an LR(1) machine} is to find a minimum merge scheme of the machine.

           \begin{table}
           \begin{center}
           \caption{The four grammars constructed from graphs with 3 nodes by our algorithm. (a) is for graphs with no edges
  (Figure \ref{G33} (c)); (b) is for graphs with one edge (Figure \ref{G33} (d)); (c) is for graphs with two edges 
  (Figure \ref{G33} (e)); and (d) is for graphs with three edges (Figure \ref{G33} (f)). 
  The production rules are classified into five categories.
  In particular, the boxed production rules in (c) are new rules added to the
  grammar on the right column in Table \ref{table2}.\label{table3}}  
           \begin{tabular}{|l|l|l|l|} \hline
  (a) & (b)  & (c)  & (d)  \\
 P ::= S\$  &  P ::= S\$ &  P ::= S\$ &  P ::= S\$ \\
   &   &   &   \\
 S ::= 1Xa  &  S ::= 1X) &  S ::= 1X) & S ::= 1X)  \\
 S ::= 1Yb  &  S ::= 1Yb &  S ::= 1Yb & S ::= 1Yb  \\
 S ::= 1Ze  &  S ::= 1Ze &  \fbox{S ::= 1Z=} & S ::= 1Z=  \\
 S ::= 1Vf  &  S ::= 1Vf &  \fbox{S ::= 1Vf} & S ::= 1Vf  \\
   &   &   &   \\
 S ::= 2Xc  &  S ::= 2Xc &  S ::= 2Xc & S ::= 2Xc  \\
 S ::= 2Yd  &  S ::= 2Y) &  S ::= 2Y) & S ::= 2Y)  \\
 S ::= 2Zg  &  S ::= 2Zg &  \fbox{S ::= 2Zg} & S ::= 2Z=  \\
 S ::= 2Vh  &  S ::= 2Vh &  \fbox{S ::= 2Vh} & S ::= 2Vh  \\
   &   &   &   \\
 S ::= 3Xi  &  S ::= 3Xi &  \fbox{S ::= 3Xi} & S ::= 3Xi  \\
 S ::= 3Yj  &  S ::= 3Yj &  \fbox{S ::= 3Yj} & S ::= 3Yj  \\
 S ::= 3Zk  &  S ::= 3Zk &  \fbox{S ::= 3Zk} & S ::= 3Zk  \\
 S ::= 3Vm  &  S ::= 3Vm &  \fbox{S ::= 3V=} & S ::= 3V=  \\ 
   &   &   &   \\
 X ::= @    &  X ::= @   &  X ::= @   & X ::= @   \\  
 Y ::= @    &  Y ::= @   &  Y ::= @   & Y ::= @   \\  
 Z ::= @    &  Z ::= @   &  \fbox{Z ::= @}   & Z ::= @   \\  
 V ::= @    &  V ::= @   &  \fbox{V ::= @}   & V ::= @   \\   \hline
           \end{tabular}
           \end{center}
           \end{table}

 We build a context-free grammar $G$ for a given color graph $F$ {\it inductively}.
 The LR(1) machine is then derived from the grammar.
 We did not construct the LR(1) machines directly because context-free grammars
 are easier to generate.

 Assume color graph $F$ has $n$ nodes.
 Then the constructed machine $M$ has $n$ states that are similar to one another;
 the remaining states are distinct and can be ignored in the discussion of merging similar states.
 There is 1-1 correspondence between the $n$ nodes in $F$ and the $n$ similar states in $M$.
 We claim that $M$ satisfies the following property:
	\begin{quote}
  {\it $F$ may be colored with $k$ colors if and only if $n-k$ pairs of
  similar states in $M$ may be merged (so that only $k$ similar states remain).}
	\end{quote}

 If there is any algorithm that can calculate the minimum LR(1) machine $M_{min}$
 from $M$ by merging certain pairs of similar states,
 we can use that algorithm to solve the node-coloring problem---for all states 
 that  are merged into a single state in $M_{min}$, their corresponding nodes 
 in $F$ have the same color.

 Due to the above property, we have successfully reduced the node-coloring problem to
 the minimization problem.
 Because the node-coloring problem is NP-hard \cite{Garey1979},
 the minimization problem is also NP-hard.

 To construct a context-free grammar from the color graph $F$, we first choose 
 two arbitrary nodes $t_{1}$ and $t_{2}$.
 There are two cases, shown in Figure \ref{G33} (a) and (b):
 there is no or one edge $t_{1}$---$t_{2}$.
 Then one of the grammars in Table \ref{table2} is selected.

 Assume that there is an edge $t_{1}$---$t_{2}$ in $F$.
 Then the  grammar on the right in Table \ref{table2} is selected.
 The corresponding LR(1) machine is shown in Figure \ref{G32-1}, in which
 there are two similar states ($s_{1}$ and $s_{2}$).
 Merging the two similar states will cause a conflict due to the terminal 
 symbol ``)''.\footnote{In our constructed grammars, the numbers, such 
       as 1, 2, 14, 27, {\it etc.}, are terminals and indicate a 
       similar state in the resulting LR(1) machine
       and the order the corresponding nodes in $F$ are chosen.
       The upper-case English letters, such as A, B, {\it etc}., denote nonterminals.
       The lower-case English letters, such as a, b, {\it etc}., denote terminals that are used only
       once in the grammar. These lower-case letters will not cause conflicts.    
       The punctuation marks, such as ``)'' and ``='', are terminals that will cause conflicts.}
 The grammar is carefully constructed so that the conflict $s_{1} \leftrightarrow s_{2}$ 
 corresponds to the edge $t_{1}$---$t_{2}$ in $F$.

 On the other hand, if $t_{1}$ and $t_{2}$ in $F$ are not connected,
 the grammar on the left in Table \ref{table2} will be selected.
 Figure \ref{G32-0} is the LR(1) machine for that grammar.
 There are two similar states in that machine ($s_{1}$ and $s_{2}$).
 The two similar states can be merged without conflicts.
 This corresponds to the fact that 
 $t_{1}$ and $t_{2}$ in Figure \ref{G33} (a) can have the same color
 since there is no edge connecting them. 

 Note that the notion of {\it ``two (similar) states can be merged''} in 
 the LR(1) machine is closely 
 related to the notion of {\it ``two nodes can have there same color''}
 due to our construction.

 The remaining nodes in $F$ are chosen one by one in an arbitrary order.
 By adding one node at a time, we can gradually construct grammars 
 $G_{2}, G_{3}, G_{4}, \ldots , G_{n}$.



  %
 %


           \begin{figure}
 \begin{flushleft}
  1. \ \ \ Nodes in graph $P$ are listed as $v_{1}, v_{2}, \ldots , v_{n}$;  \\
  2. \ \ \ {\bf if} there is an edge $v_{1}$---$v_{2}$ {\bf then} G := the left grammar in Table \ref{table2} ;  \\
 3. \ \ \ {\bf else} G := the right grammar in Table \ref{table2};  \\
 4. \ \ \ NewNonTerm := \{ X, Y \}; \\
 5. \ \ \ {\bf for} $\mu$ := 3 {\bf to} n {\bf do} \\
 6. \ \ \ \ \ \ generate two new nonterminals, called $\Delta$ and $\Theta$; \\
 7. \ \ \ \ \ \ generate two new terminals, called $\phi$ and $\omega$; \\
 8. \ \ \ \ \ \ generate four production rules:  \\
 9. \ \ \ \ \ \ \ \ \ (``S ::= '' $\mu$ $\Delta$ $\phi$) and (``S ::= '' $\mu$ $\Theta$ $\omega$) and \\
 10. \ \ \ \ \ \ \ \ \ ($\Delta$ `` ::= @'') and ($\Theta$ `` ::= @''); \\
 11. \ \ \ \ \ \ {\bf for} each nonterminal $\Pi \in $ NewNonTerm {\bf do}  \\
 12. \ \ \ \ \ \ \ \ \ generate a new terminal, called $\psi$; \\
 13. \ \ \ \ \ \ \ \ \ generate a production rule: (``S ::= '' $\mu$ $\Pi$ $\psi$); \\
 14. \ \ \ \ \ \ {\bf end}; \\
 15. \ \ \ \ \ \ {\bf for} $\delta$ := 1 {\bf to} $\mu - 1$ {\bf do} \\
 16. \ \ \ \ \ \ \ \ \ generate two new terminals, called $\tau$ and $\rho$; \\
 17. \ \ \ \ \ \ \ \ \ generate a production rule: (``S ::= '' $\delta$ $\Theta$ $\rho$); \\
 18. \ \ \ \ \ \ \ \ \ {\bf if} there is an edge $v_{\mu} \leftrightarrow v_{\delta}$ {\bf then} \\
 19. \ \ \ \ \ \ \ \ \ \ \ \ /* {\it The following rule will cause a conflict due to $\omega$.} */ \\
 20. \ \ \ \ \ \ \ \ \ \ \ \ generate a production rule: (``S ::= '' $\delta$ $\Delta$ $\omega$); \\
 21. \ \ \ \ \ \ \ \ \ {\bf else} /* {\it The following rule will NOT cause a conflict.} */  \\
 22. \ \ \ \ \ \ \ \ \ \ \ \ generate a production rule: (``S ::= '' $\delta$ $\Delta$ $\tau$); \\
 23. \ \ \ \ \ \ {\bf end}; \\
 24. \ \ \ \ \ \ NewNonTerm := NewNonTerm $\cup \ \{ \Delta, \Theta \}$; \\
 25. \ \ \ {\bf end}; \\ 
           \caption{Algorithm for generating a context-free grammar from a graph.
    The four rules generated at lines 9, 17, and 20 will cause a conflict in the LR(1) machine
    due to the terminal $\omega$. \label{list1}}  
    \end{flushleft}
           \end{figure}

 \subsection{Extending Grammar $G_{\mu-1}$ to Grammar $G_{\mu}$}

 In this subsection, we explain the steps for extending grammar $G_{\mu-1}$ to grammar $G_{\mu}$.
 The complete algorithm for generating a context-free grammar from a graph is shown in Figure \ref{list1}.

           \begin{figure}
           \begin{center}
            \resizebox{4.5in}{!}{\includegraphics{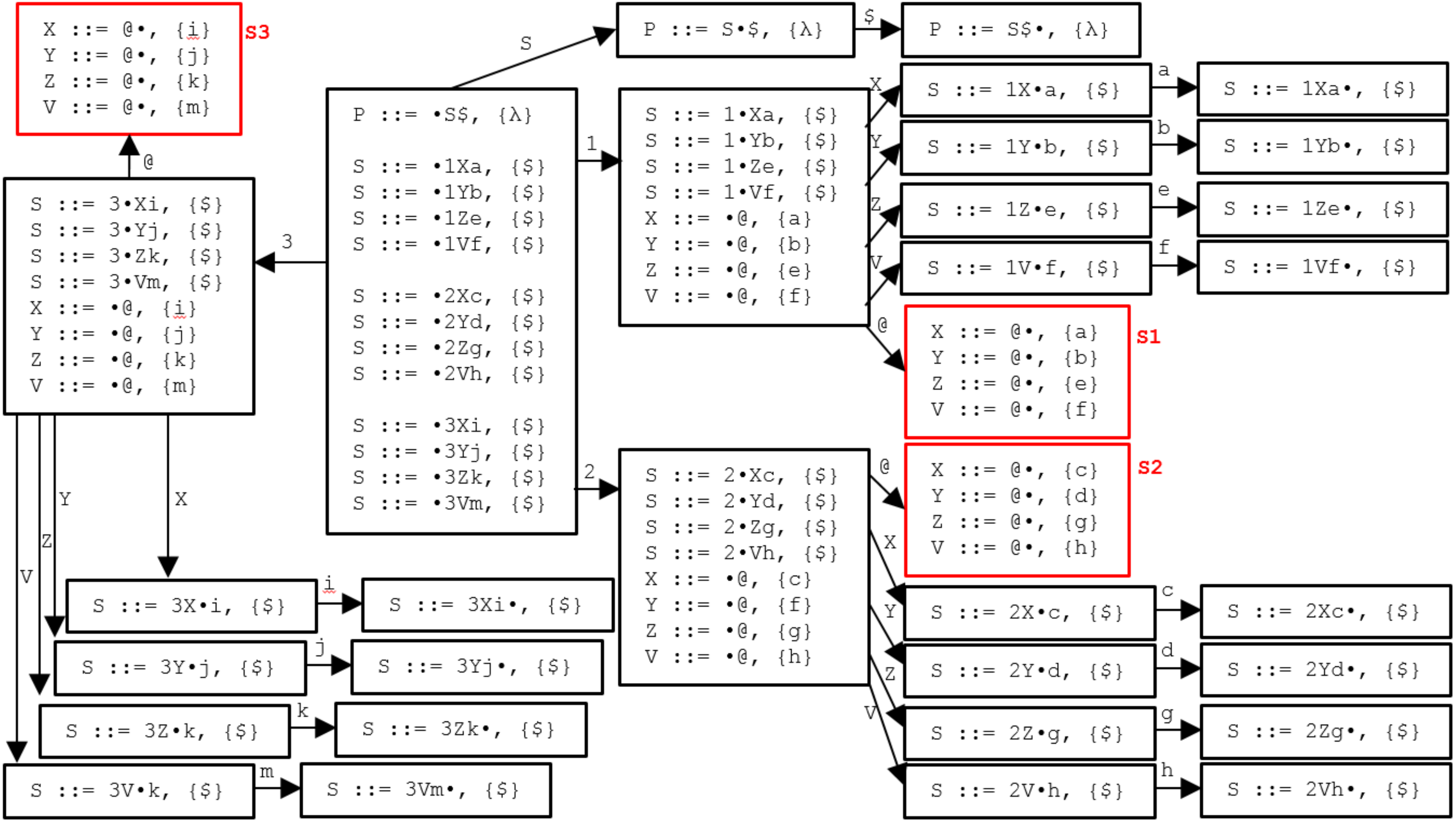}}
            \caption{The LR(1) machine for the graph in Figure \ref{G33} (c).    \label{G31-0}}
            \end{center}
           \end{figure}

           \begin{figure}
           \begin{center}
            \resizebox{4.5in}{!}{\includegraphics{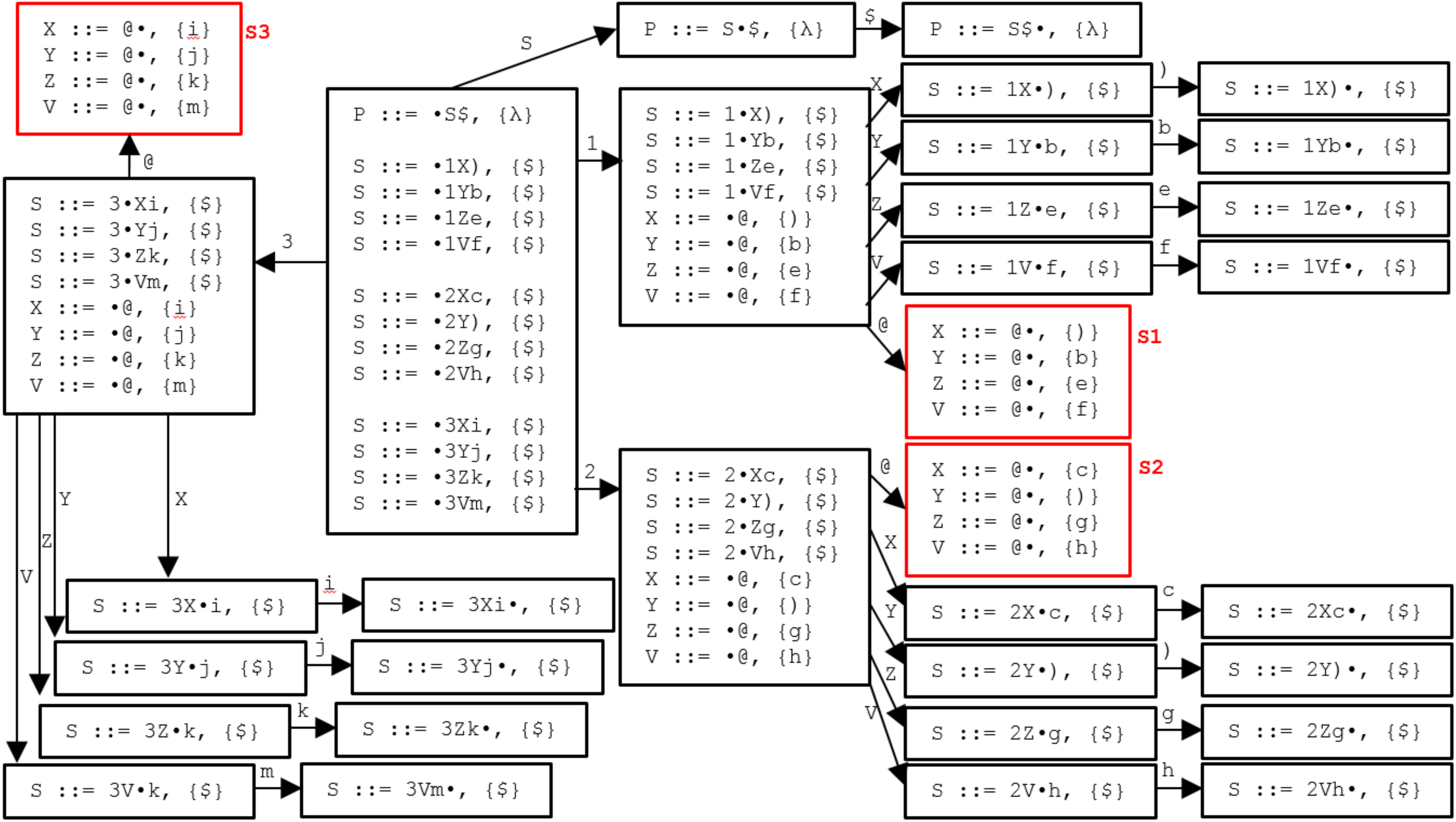}}
            \caption{The LR(1) machine for the graph in Figure \ref{G33} (d).\label{G31-1}}  
            \end{center}
           \end{figure}

 We may list the nodes in a color graph $F$ as $v_{1}, v_{2}, \ldots , v_{n}$.
 We first consider the two nodes $v_{1}$ and $v_{2}$.
 One of the grammars in Table \ref{table2} is selected according to whether 
 there is an edge $v_{1}$---$t_{2}$.
 Call the resulting grammar $G_{2}$.
 
 The remaining nodes $v_{3}, v_{4}, \ldots , v_{n}$ are added one at a time, resulting
 in grammars $G_{3}, G_{4}, \ldots , G_{n}$.
 (This is done in the {\tt for}-loop between lines 5 and 25 in Figure \ref{list1}.)
 Each $G_{\mu}$ is constructed for the subgraph of $F$ consisting of nodes $v_{1}, v_{2}, \ldots , v_{\mu}$
 and all edges incident on these nodes.
 The last grammar $G_{n}$ is the intended grammar.

 Grammar $G_{\mu}$ is obtained from the previous grammar $G_{\mu-1}$ by adding two new nonterminals $\Delta$ and $\Theta$,
 $4\mu-3$ new terminals ($\mu$, $\phi$, $\omega$, $\ldots$), and $4\mu-2$ new production rules.

 Consider node $v_{\mu}$.
 Firstly, four new production rules (S ::= $\mu$ $\Delta$ $\phi$), \fbox{(S ::= $\mu$ $\Theta$ $\omega$)}, 
 ($\Delta$ ::= @), and ($\Theta$ ::= @), are generated, where S is the nonterminal taken from the initial grammar $G_{2}$,
 $\mu$ is the index of
 the current iteration that is treated as an integer terminal,
 $\Delta$ and $\Theta$ are two newly created nonterminals, @ is the terminal taken from the initial grammar $G_{2}$,
 and $\phi$ and $\omega$ are two of the new terminals.
 (We will use the numbers 1, 2, 3, {\it etc.}, to denote the new terminals in the order of creation, 
 as shown in grammars in Table \ref{table3}.
 Each number also corresponds to a new similar state in the final LR(1) machine.)



 The variable NewNonTerm is the set of nonterminals that are created.
 At the beginning of the ``$\mu=k$'' iteration of the {\tt for}-loop, there are $2k -4$ nonterminals
 in the variable NewNonTerm.
 During each iteration, we will create one new production rule (S ::= $\mu$ $\Pi$ $\psi$) for every nonterminal
 $\Pi \in $ NewNonterm, where $\mu$ is the loop index and $\psi$ is a new terminal used only in this production rule.

           \begin{figure}
           \begin{center}
            \resizebox{4.5in}{!}{\includegraphics{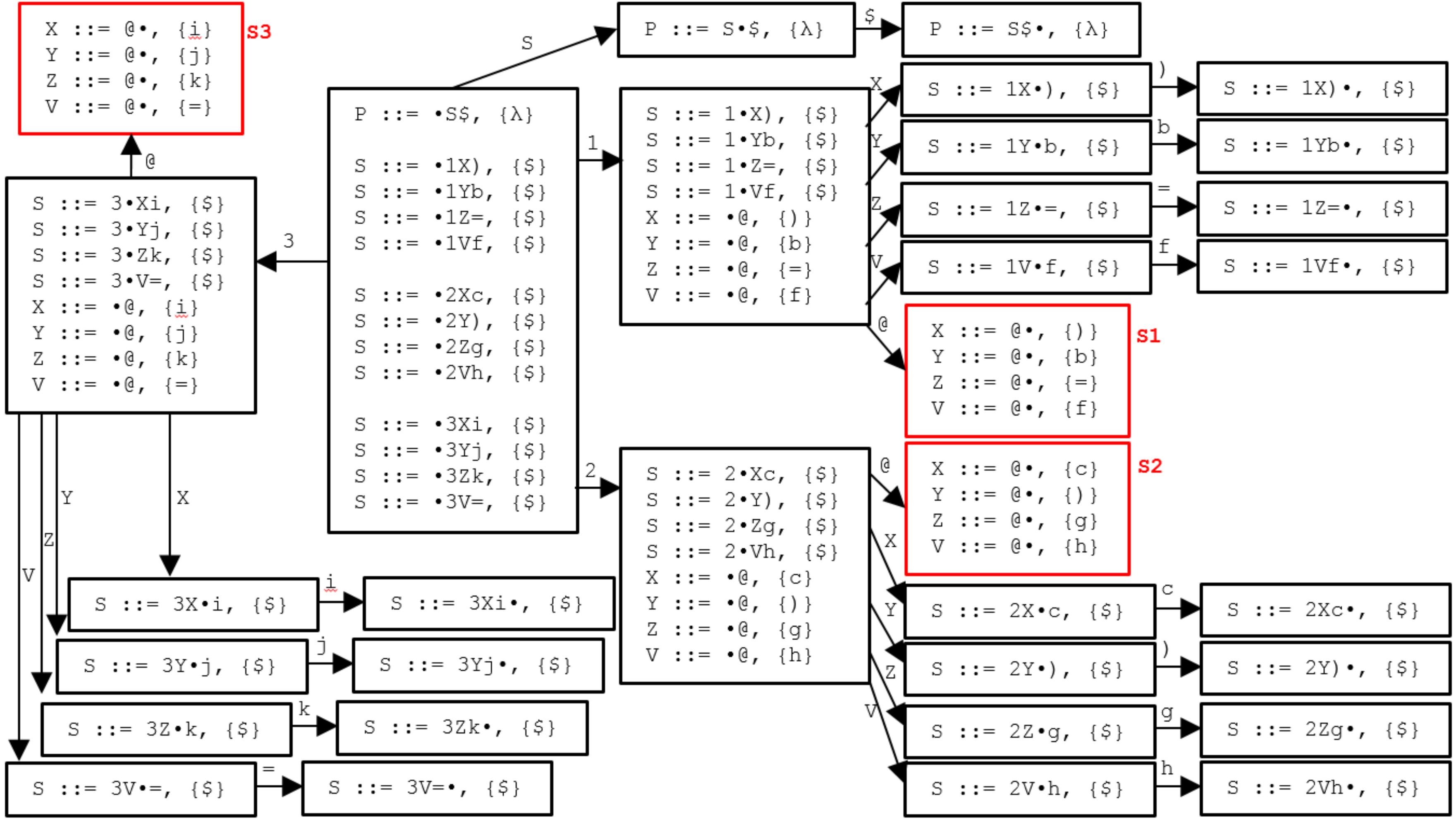}}
            \caption{The LR(1) machine for the graph in Figure \ref{G33} (e). \label{G31-2}}  
            \end{center}
           \end{figure}

 Consider node $v_{\mu}$.
 $v_{\mu}$ might be connected to some of $v_{1}, v_{2}, \ldots , v_{\mu-1}$.
 Let $U = \{ \delta \ | \ 1 \le \delta < \mu, \mbox{there is an edge } v_{\delta}$---$v_{\mu} \}$ and
     $W = \{ \delta \ | \ 1 \le \delta < \mu, \delta \not\in U \} $.
 For each $\delta \in U$, we generate two new production rules:
 {(S ::= $\delta$ $\Theta$ $\rho$)} and \fbox{(S ::= $\delta$ $\Delta$ $\omega$)},
 where $\rho$ is a new terminal and $\omega$ is a terminal generated at the beginning of the current iteration.
 For each $\delta \in W$, we generate two new production rules:
 {(S ::= $\delta$ $\Theta$ $\rho$)} and {(S ::= $\delta$ $\Delta$ $\tau$)},
 where $\rho$ and $\tau$ are two new terminals.
 (This is done in the {\tt for} loop at line 15.)

           \begin{figure}
           \begin{center}
            \resizebox{4.5in}{!}{\includegraphics{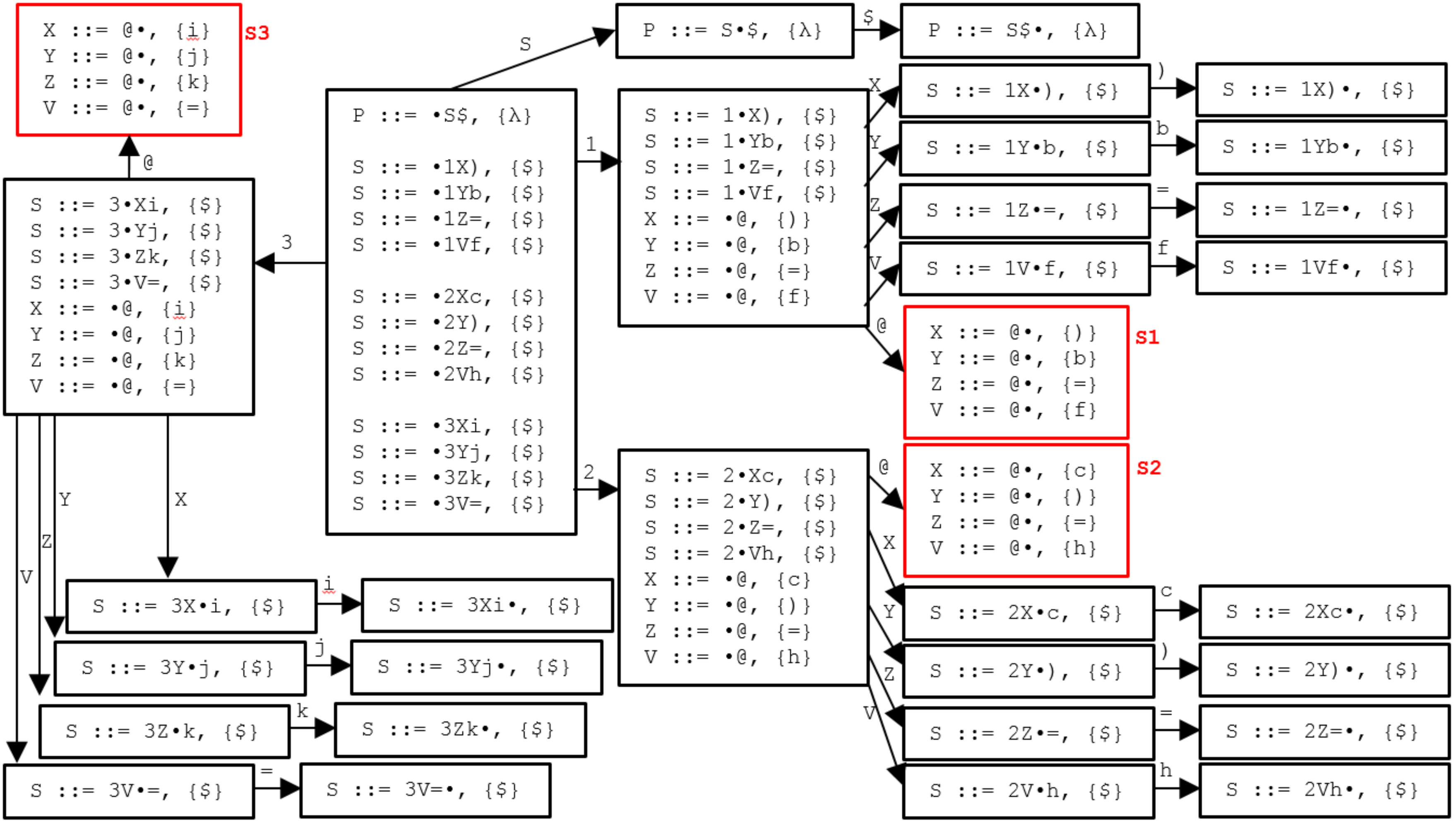}}
            \caption{The LR(1) machine for the graph in Figure \ref{G33} (f).  USELESS G31-3 (Figure 7) \label{G31-3}} 
            \end{center}
           \end{figure}

 Note that every production rule of the form (S ::= $ii$ $nn$ $tt$) will not cause a 
 conflict (in the LR(1) machine for this grammar) as long as the terminal $tt$ is not used 
 in any other production rules.
 (An example is the production rule (S ::= 2 X c) in the two grammars in Table \ref{table2}.)
 Therefore, only the two production rules \fbox{(S ::= $\mu$ $\Theta$ $\omega$)}
 and \fbox{(S ::= $\delta$ $\Delta$ $\omega$)}
 will cause a conflict (together with other production rules).

 Note also that different terminals are generated during different iterations of the {\tt for} loop at line 5.
 Therefore, production rules generated in different iterations will not conflict with each other.

 {\it Example}.
 Suppose the color graph in Figure \ref{G33} (b) is extended to that in Figure (e).
 The corresponding grammars are shown on the right column in Table \ref{table2} and 
 on the third column in Table \ref{table3}, respectively.
 Note that the latter grammar is extended from the former grammar.
 The boxed production rules in the latter grammar are added by the algorithm in Figure \ref{list1}.

 {\it Example}.
 Figures \ref{G31-0}, \ref{G31-1}, \ref{G31-2}, and \ref{G31-3} are the constructed state machines for the graphs in 
 Figure \ref{G33} (c), (d), (e), and (f), respectively.
 $\Box$

 The grammar on the third column in Table \ref{table3} is a typical grammar generated by our algorithm.
 The production rules are classified into five categories:
		\begin{enumerate}
		\item one starting production rule ({\it i.e.}, P ::= S \$)
		\item four production rules of the form ($\Pi$ ::= @)
		\item four production rules whose right-hand sides begin with the terminal 1
		\item four production rules whose right-hand sides begin with the terminal 2
		\item four production rules whose right-hand sides begin with the terminal 3
		\end{enumerate}

 Now consider the constructed LR(1) machine in Figure \ref{G31-2}.
 Note that all items derived from rules of categories 1, 3, 4, and 5, appear only once in the whole LR(1) machine.
 Any state containing any of these items will not be similar to any other state and hence can be ignored.
 We could focus on states consisting solely of items derived from production rules of category 2.
 There are only 3 such states, which are indeed similar to one another.
 Each such state has all items of the form ($\Pi$ ::= @ $\bullet$, $\ldots$), where $\Pi$ is a nonterminal except P and S.

 Another characteristic of the constructed LR(1) machine in Figure \ref{G31-2} is that there are no cycles.
 The longest path contains 3 steps.

 In fact, all grammars generated by the algorithm in Figure \ref{list1} share the above characteristics. 
 They help us to infer properties of the minimized LR(1) machines.

 %
 %
 %
 %
 %

 In the corresponding state machine in Figure \ref{G31-2}, consider the four states that
 come immediately after the initial state.
 Items derived from rules in categories (3), (4) and (5) are cleanly separated because of the first symbols (which
 are integer terminals) on the right-hand sides of the rules.
 Hence, except the four states that come immediately after the initial state, those items whose first symbols 
 on the right-hand sides are different will never be mixed in the same state in the state machine.
 Items derived from rules in category (2) are quite similar---actually all items of the 
 form ($\Pi$ ::= $\bullet$ @, where is $\Pi$ is a nonterminal) appear in every state that comes immediately 
 after the initial state (we will ignore the
 starting production rule in this discussion).
 Furthermore, all items of the form ($\Pi$ ::= @ $\bullet$) appear in states that come two steps after the initial state.
 It is these states (which contain all items of the form ($\Pi$ ::= @ $\bullet$) and no other items) 
 that are similar to one another.
 All other states are not similar to any other states and hence can be ignored when we discuss the merging
 of similar states.

 Therefore, we can create or avoid conflicts among similar states by carefully adjusting the last terminal $\psi$ in
 rules of the form (S ::= $\mu$ $\Pi$ $\psi$), where $\mu$ is an integer terminal,
 $\Pi$ is a nonterminal, and $\psi$ is a terminal.
 In the four grammars in Table \ref{table3}, when $\psi$ is a lower-case English letter ({\it e.g.}, ``b'' or ``i''),
 that rule will not cause any
 conflict because the lower-case letter appears only once in the whole grammar.  
 On the other hand, when $\psi$ is a punctuation marks ({\it e.g.}, ``)'' or ``=''), a conflict is intentionally
 added to the grammar because that punctuation mark is used in two different production rules.
 The above discussion is related lines 18-22 in the algorithm in Figure \ref{list1}.

 \subsection{The Constructed Grammar and LR(1) Machine}

 In general there are $2n$ nonterminals, $2n^{2}-n+2-e$ terminals, and $2n^{2}-1$ production rules 
 in the final constructed grammar $G_{n}$,
 where $n$ is the number of nodes and $e$ is the number of edges in the original color graph.
 All nonterminals except P and S are called {\it NewNonTerm}s (new nonterminals).
 There are $2n-2$ NewNonTerms.
 There are one starting rule and $2n-2$ rules of the form ($\Pi$ ::= @), one for each NewNonTerm $\Pi$ .

 The remaining $n(2n-2)$ rules have the form (S ::= $\mu$ $\Pi$ $\phi$), where $S$ is the nonterminal taken from 
 a grammar in Table \ref{table2},
 $\mu$ is an integer terminal, 
 $1 \le \mu \le n$, $\Pi$ is a nonterminal, and $\phi$ is a terminal.
 These production rules are divided into $n$ groups according to the first (integer terminal) symbols on the right-hand sides.

 Each group consists of exactly $2n-2$ rules.  
 In each group, each of the $2{n}$ nonterminals, except P and S (which are taken from a grammar in Table \ref{table2})
 appears as the second (nonterminal) symbol on the right-hand side of a production rule.

 There are $4n^{2}-2n+3$ states in the LR(1) machine $M_{n}$ derived from the constructed grammar $G_{n}$.
 One is the starting state.
 There are $n+1$ successor states immediately following the starting state.
 One successor state contains the item (P ::= S $\bullet$ \$, $\{ \lambda \}$).
 Each of the $n$ remaining successor states consists of $2n-2$ items (which belong to the same group of production rules
 defined above) of the form 
 (S ::= $\mu$ $\bullet$ $\Pi$ $\phi$, $\{$ \$ $\}$) and $2n-2$ items of the form ($\Pi$ ::= $\bullet$ @, $\ldots$).
 Then each successor state is, in turn, followed by another state, denoted as $s_{\mu}$, of $2n-2$ items of 
 the form ($\Pi$ ::= @ $\bullet$, $\ldots$).
 
 In the whole LR(1) machine, there are $n$ such states ({\it i.e.}, $s_{\mu}$, where $\mu = 1, 2, \ldots, n$)
 because there are $n$ groups of production rules.
 These $s_{\mu}$ states are similar to one another.
 State $s_{\mu}$ corresponds to node $v_{\mu}$ in the original color graph, where $\mu = 1, 2, \ldots, n$.
 All other states in the derived LR(1) machine are dissimilar to one another.  
 They can be ignored when we attempt to merge {\it similar} states.

 After the last grammar $G_{n}$ is constructed, the corresponding LR(1) state machine $M$ is derived from the grammar.
 Note that the nonterminal P never appears in the right-hand side of a production rule.
 The nonterminal S appears only once in the right-hand side of only one production.
 Thus, P and S are never involved in a conflict.
 A conflict in the LR(1) machine, if it ever occurs, must occur due to two items of the form
 ($\Theta$ ::= @ $\bullet$, $\{ \omega, \ldots \}$) and ($\Delta$ ::= @ $\bullet$, $\{ \omega, \ldots \}$),
 where $\Theta$ and $\Delta$ are two different NewNonTerms and $\omega$ is a terminal.

 Note that $\omega$ is a terminal that can follow $\Theta$ and $\Delta$ in some production rules.
 However, since $\Theta$ and $\Delta$ are NewNonTerms, they can appear only in production rules of
 the form (S ::= $\mu$ $\Theta$ $\omega$) and  (S ::= $\delta$ $\Delta$ $\omega$).
 The two production rules are carefully planted in the grammar (see lines 9 and 20 in the algorithm in Figure \ref{list1}).

 We will use an algorithm to calculate the minimum state machine from $M$.   

 \subsection{Recovering a Minimum Node-Coloring from a Minimum LR(1) Machine}

 In this subsection, we will describe how to recover a minimum coloring of the original  color graph from the
 minimum state machine.

\begin{figure}
           \begin{center}
            \resizebox{4.5in}{!}{\includegraphics{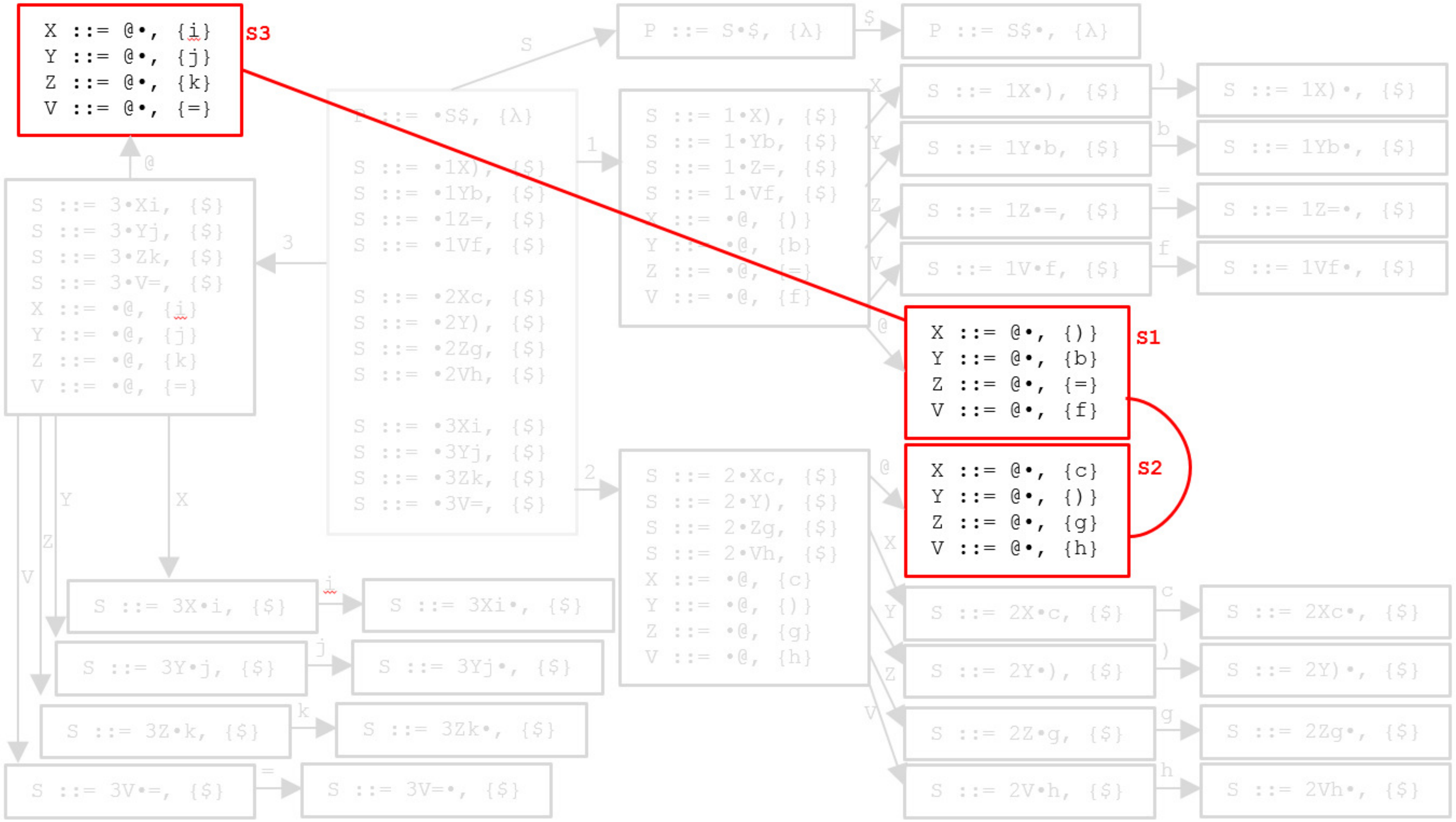}}
            \caption{The conflict graph. After removing the states that are not similar to any other states, 
                only three states are left. We may add an edge $s_{1} \leftrightarrow s_{2}$
                to indicate there is a conflict edge $s_{1} \leftrightarrow s_{2}$
                and another conflict edge $s_{1} \leftrightarrow s_{3}$.  \label{G31-2-reduced}}
            \end{center}
           \end{figure}

 For the purpose of merging similar states, we may ignore all states that are not similar to any other states.
 To make conflicts among states explicit we add a {\it conflict edge} between two states 
 if a conflict will occur when the two states are merged.  
 The state machine in Figure \ref{G31-2} becomes Figure \ref{G31-2-reduced}, which is called a {\it conflict graph}.
 Due to our construction, all states in the conflict graph (Figure \ref{G31-2-reduced}) are similar to one another.
 In Figure \ref{G31-2-reduced}, there are three (similar) states and two conflict edges $s_{1} \leftrightarrow s_{2}$
                and $s_{1} \leftrightarrow s_{3}$.

 Finding a minimum merge scheme for the LR(1) machine is identical to finding a minimum merge scheme for the conflict graph.
 So we will focus on the conflict graph instead.


 The reader may find that the conflict graph (Figure \ref{G31-2-reduced}) is isomorphic to the original color graph
 (Figure \ref{G33} (e)).
 This is due to the construction of the context-free grammar.

 For the former (the conflict graph) we wish to merge as many states as possible under the constraint that two states 
 connected by an edge, such as $s_{1} \leftrightarrow s_{2}$, should not be merged.
 For the latter (the color graph), we wish to assign colors to nodes under the constraint that two
 nodes connected by an edge, such as $t_{1} \leftrightarrow t_{2}$, should not be assigned the same color.
 Since the conflict graph and the color graph are isomorphic, it is easy to see that
 every coloring of the color graph with $k$ colors corresponds to a merge scheme of 
 the conflict graph with $k$ partition blocks, and {\it vice versa}.
 A detailed argument is in the next paragraph.

 Suppose in a coloring of the color graph, nodes $t_{i}$ and $t_{j}$ are
 assigned the same color.
 Due to the rule of coloring, $t_{i}$ and $t_{j}$ are not connected by an edge.
 The corresponding states $s_{i}$ and $s_{j}$ in the conflict graph
 are not connected by an edge either due to the construction of the grammar.
 Therefore, states $s_{i}$ and $s_{j}$ can be merged together.
 Hence, a coloring of the color graph correspond to a merge scheme of the conflict graph.
 By the same argument, a merge scheme of the conflict graph
 correspond to a coloring of the color graph.

 \subsection{Time Analysis of the Reduction}

 Constructing a context-free grammar from a graph (the algorithm in Figure \ref{list1}) takes polynomial time.
 Note that the constructed grammar does not contain recursive rules, that is, there is no derivation of
 the form $A \Rightarrow^{*} \ldots A \ldots$.
 Deriving the LR(1) state machine from such a grammar also takes polynomial time because there are
   $4n^{2}-2n+3$ states in the derived LR(1) machine.
 Constructing the conflict graph from the LR(1) machine also takes polynomial time.
 Therefore the reduction takes polynomial time.
 
 \section{Conclusion}  

 We have reduced the node-coloring problem to the minimization problem of the LR(1) state machines.
 Therefore, the minimization problem is NP-hard.

 There are efficient algorithms for minimization of finite state machines.
 LR(0) state machines are minimum by its construction.
 We show that LR(1) state machines cannot be easily minimized in general.

 Note that minimizing an LR(1) machine is quite different from minimizing a general finite state machine.
 For one thing, we need to examine the items in the states of an LR(1) machine.
 On the other hand, minimizing a general finite state machine does not consider the ``contents'' of the states.

  \ \\
  \ \\

 \input{ref}


 \newpage
 {\bf Appendix 1} A further example.

\begin{figure}
           \begin{center}
            \resizebox{1.5in}{!}{\includegraphics{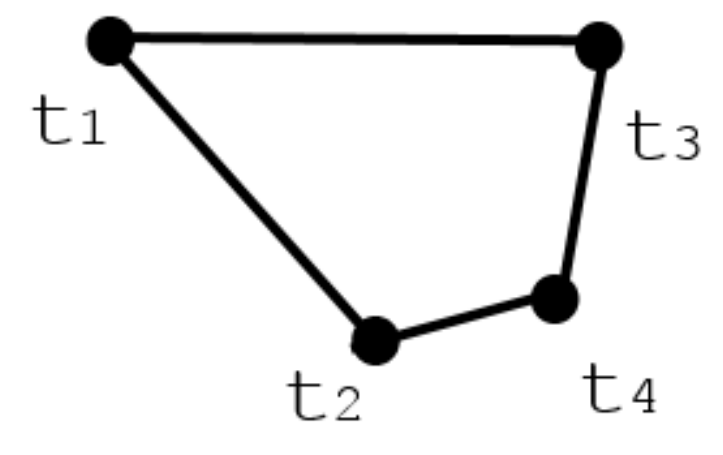}}
            \caption{The 4-node graph for coloring. \label{G55}}  
            \end{center}
           \end{figure}

 Suppose we add a new node $t_{4}$ and two new edges $t_{2}$---$t_{4}$
 and $t_{3}$---$t_{4}$ to Figure \ref{G33} (e).
 The resulting graph is shown in Figure \ref{G55}.
 The corresponding extended grammar (which is extended from the one in Table \ref{table3} (c)) is shown below. 
 The LR(1) machine is shown in Figure \ref{G34-4-new}.
 \begin{center}
 \begin{tabular}{lllll} 
 P ::= S\$ &  &  &  &  \\ 
 S ::= 1X) & S ::= 2Xc & S ::= 3Xi & S ::= 4Xr & X ::= @ \\ 
 S ::= 1Yb & S ::= 2Y) & S ::= 3Yj & S ::= 4Ys & Y ::= @ \\ 
 S ::= 1Z= & S ::= 2Zg & S ::= 3Zk & S ::= 4Zt & Z ::= @ \\ 
 S ::= 1Vf & S ::= 2Vh & S ::= 3V= & S ::= 4Vq & V ::= @ \\ 
 S ::= 1Qm & S ::= 2Q\#& S ::= 3Q\#& S ::= 4Qv & Q ::= @ \\ 
 S ::= 1Rn & S ::= 2Rp & S ::= 3Ru & S ::= 4R\#& R ::= @ \\ 
 \end{tabular}
 \end{center}
 
           \begin{figure}
           \begin{center}
            \resizebox{4.5in}{!}{\includegraphics{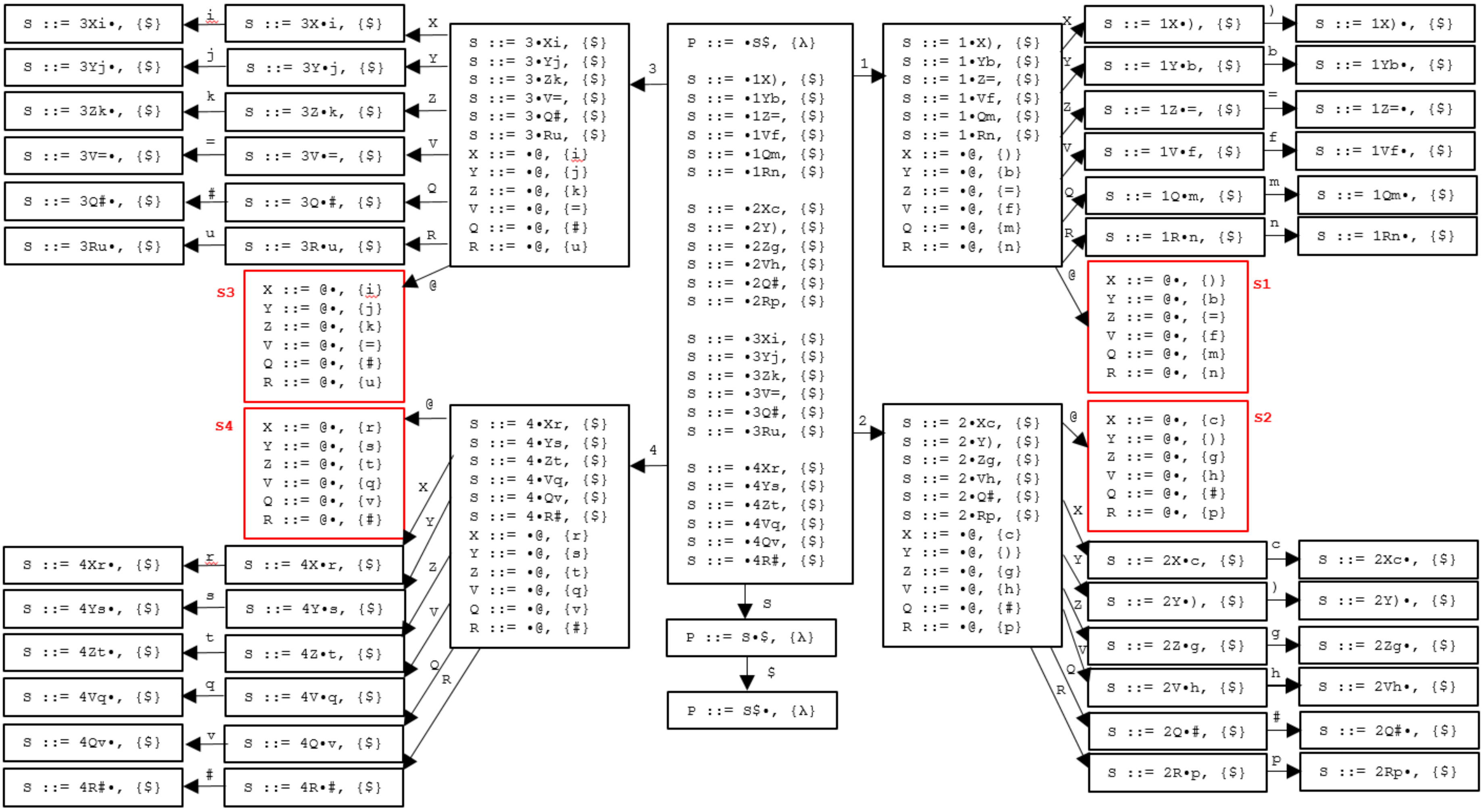}}
            \caption{The LR(1) machine for the graph in Figure \ref{G55}.
             The four red states $s_{1}, s_{2}, s_{3}, s_{4}$ correspond to the four nodes in Figure \ref{G55}. 
             Other nodes are not similar to any other nodes and hence are not involved in merging.
             \label{G34-4-new}}  
            \end{center}
           \end{figure}
 \  \\

 {\bf Appendix 2}.
 The sizes of the LR(1) and LALR(1) machines for several practical programming languages are shown in Table \ref{tb20}, 
 which is adapted from \cite{Chen2011}.

           \begin{table}[t]
           \begin {center}
           \caption{Sizes of LR(1) and LALR(1) state machines} \label{tb20}
           \begin{tabular}{c r r r} \hline
            grammar & LR(1)  & LALR(1) & ratio \\ \hline
            Ada  & 12786 &  861 & 14.9 \\
            C    & 1572  &  350 & 4.5  \\
            C++ 5.0  & 9785 &  1257 & 7.8\\
            Java 1.1  & 2479 & 429  & 5.8 \\
            Pascal    & 2245 & 413  & 5.4 \\  \hline
           \end{tabular}
           \end {center}
           \end{table}

\newpage
 \  \\
 \  \\
 \  \\
\newpage
 {\bf Appendix 3}.

Additional state machines.

\begin{figure}
           \begin{center}
            \resizebox{4.5in}{!}{\includegraphics{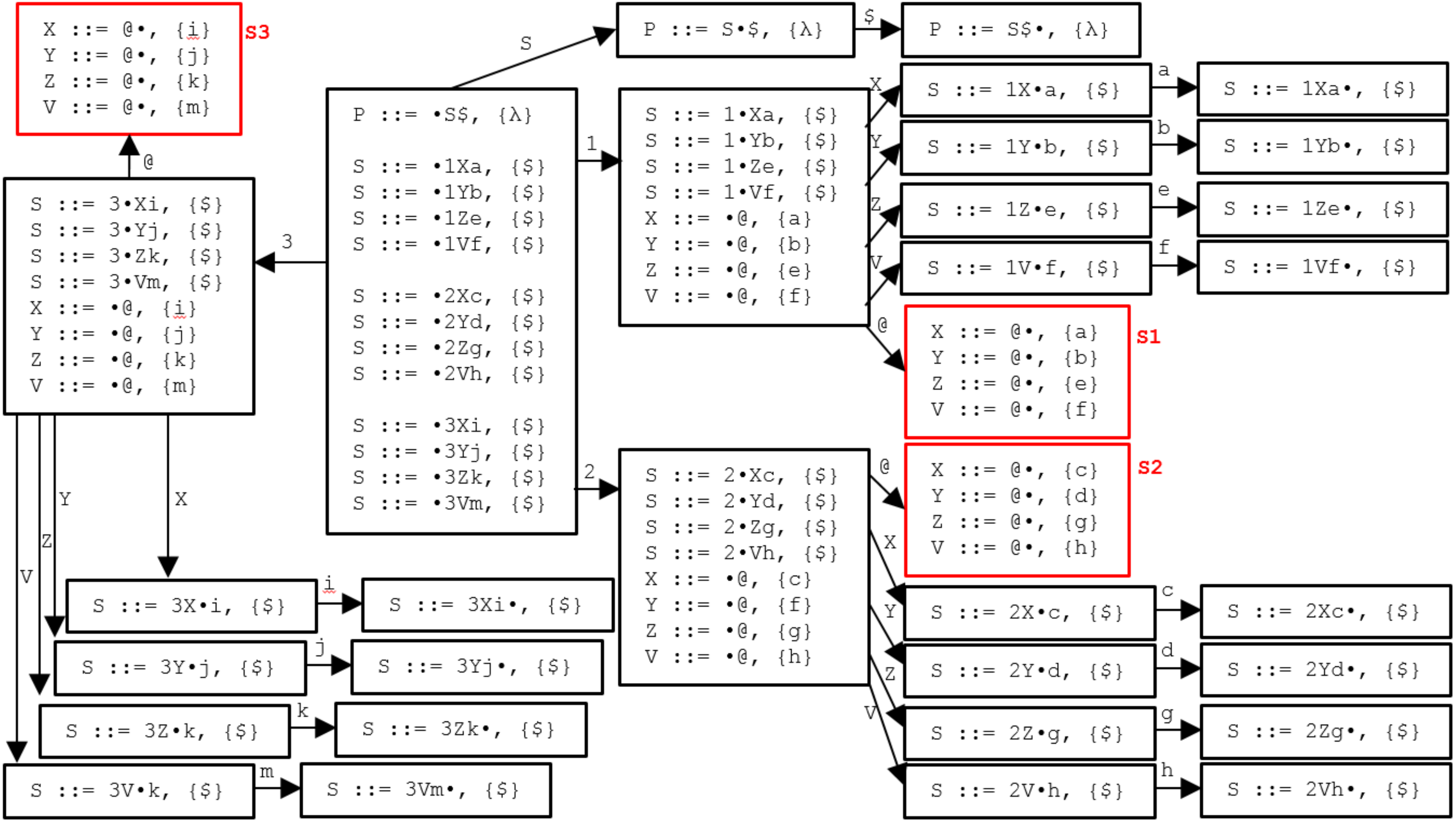}}
            \caption{The LR(1) machine for the graph in Figure \ref{G33} (c).    \label{G31-000}}
            \end{center}
           \end{figure}

  It seems the following picture Figure \ref{G31-000} is not referenced in the paper.

 \  \\
 \  \\
 \  \\
 {\bf Appendix 4}.

 A simple statistics.

 \begin{center}
 \begin{tabular}{c|c|c|c}  \hline
   & \# nonterminals & \# terminals & \# production rules   \\  \hline
 total for $\mu = 2$ & 4  &  8 &   7 \\ 
 total for $\mu = 3$ & 6  &  17  &   17   \\ 
 total for $\mu = 4$ & 8  & 30  &  31  \\ 
 total for $\mu = n$ & $2n$  & $2n^{2}-n+2$  &  $2n^{2}-1$  \\ \hline
 \end{tabular}
 \end{center}

 \begin{center}
 \begin{tabular}{c|c|c|c}  \hline
   & \# nonterminals & \# terminals & \# production rules   \\  \hline
 total for $\mu = 2$ & 4  &  8 &   7 \\ 
 increment for $\mu = 3$ & 2  &  9  &   10   \\ 
 increment for $\mu = 4$ & 2  & 13  &  14  \\ \hline
 increment for $\mu = n$ & 2  & $4\mu-3$  &  $4\mu-2$  \\ \hline
 total for $\mu = n$ & $2n$  & $2n^{2}-n+2$  &  $2n^{2}-1$  \\ \hline
 \end{tabular}
 \end{center}

 \  \\
 \  \\
 \  \\


 \long\def\comment#1{}
 
 \comment{

 \newpage
 Additional Stuff.

--


 [pager1977] and [IELR2009] are most closely related work.
 Both algorithms merge states when they are generated.
 In contrast, our algorithm merges states after the whole LR(1) machine has been generated.
 Our algorithm needs more space.

\ \

 Both (pager1977) and (IELR(1)) attempts to find a {\it minimal} machine.
 However, {\it minimal} simply means ``very small'' or ``locally minimum''
 rather than ``globally minimum''.(IELR(1))

\ \

 As the minimization problem is proven NP-hard, the two algorithms can only be
 {\it practical} algorithms for real-world parser generators.
 They cannot guarantee to produce the minimum state machines.

\ \


 Furthermore, the efficiency for algorithms/programs for NP-hard problems depends on many factors, including
 test cases, programming expertise, clever data structures, {\it etc}., which are not
 the merits of the algorithms themselves.
 It is hard to say which algorithms produce the smallest state machines in general.

\ \

 By the way, (pager1977) and (IELR(1)) can handle some non-LR(1) grammars in addition to
 LR(1) grammars. For a non-LR(1) grammar $G$ together with a ``disambiguation'' function $\Delta$, their algorithms
 produce an LR(1)-like state machine $M$.
 On the other hand, $G$ together with $\Delta$ corresponds
 to a deterministic PDA, which is, in turn, equivalent to an LR(1) grammar $G^{\prime}$.
 From $G^{\prime}$ there is a corresponding minimum $M^{\prime}$.
		\[ G + \Delta \Rightarrow M  \Rightarrow G^{\prime} \Rightarrow  M^{\prime}  \]
 It is unclear whether $M = M^{\prime}$.



 Many research focuses on generating LALR(1) state machines efficiently, not on
 minimizing the state machines.



 (p613-korenjak.pdf)  \\
 (p613-korenjak.pdf)

\ \

J.E. Denny and B.A. Malloy,
the IELR(1) algorithm for generating minimal LR(1) parser tables for non-LR(1) grammars with conflict resolution,
Science of Computer Programming vol 75 no 11, November 2010, pp. 943-979.
doi:10.1016/j.scico.2009.08.001 \\
 The IELR(1) algorithm (Denny2009) attempts to find the ``minimal LR(1) machine''.
 However, ``minimal'' simply means ``very small'' in the algorithm.
 This is very

\ \

 (How much bigger can an LR(1) automaton for a language be than the corresponding LR(0) automaton.pdf) How much bigger can an LR(1) automaton for a language be than the corresponding LR(0) automaton.pdf.  \\
 (How much bigger...) The size of LR(1) machines is usually 2-4x larger than
 the corresponding LR(0) machines though theoretically, it could be $2^{n}$ times larger.

\ \

 (LRParsing.pdf) Chapter 4 LR Parsing. \\
 (LRParsing.pdf) This document seems the classnotes based on Fischer's textbook.
 Nothing new to me.

\ \



 Our algorithm first builds the whole (big) LR(1) machines and tries to merge similar states.

 There are other algorithms, such as (Pager1977), which tries to build LR(1) machines.
 However, whenever a new state is generated, the algorithm attempts to merge the new state
 with an existing states, if the states satisfy {\it strong compatibility} or {\it weak compatibility}.
 Strong compatibility is more effective than the weak one.
 However, it is still not the exact condition.
 This means states not satisfying the strong compatibility could still be merged.
 In contrast our algorithm simply to merge similar states, which are already generated,
 in the reverse direction of state transition.
 In this sense our algorithm can possibly merge more states.
 The other factor that affects the merging is the order of merging for states that are not connected by a directed path.
 This is true for both our algorithm and the one in (Pager1977).

 If the input grammar is LALR(1), both (Pager1977) and our algorithm will produce an LALR(1) machine.
 On the other hand, if the grammar is LR(1) but not LALR(1), it is not clear which algorithm will produce a
 smaller machine.

 However, since minimizing LR(1) machines
 is NP-hard, all these algorithms are considered ``heuristics''.
 There is no clear evidence which algorithms will produces the smaller LR(1) machines.

 Still others simply build LR(0) machines and then add the lookahead information, resulting
 in LALR(1) machines.

 (Pager1977) can only handle LR(k) grammars.
 (Pager lane tracing algorithm) determines the required length $k$ and, if necessary, split states
 so as to remove the conflict.



 (Pager1977) D. Pager, A practical general method for constructing LR(k) parsers,
 Acta Inform. 7 (3) (1977) 249–268.\\
 (Pager1977) makes use of the usual LR(1) construction algorithm.
 When a new state is generated, the algorithm attempts to merge the new state
 with an existing state, that is, combining ``compatible'' states as they are generated.
 In my paper, states are merged in the reverse direction of the state transition in the
 state machine. In contrast, (Pager1977) need to merge states in the direction of the
 sate transition. Therefore, (Pager1977) needs to foresee the conflicts that may occur
 in the future. Therefore, he defined the notion of ``compatible states''.
 In contrast, our algorithm is much easier to understand---simply merging similar states and check if there are conflicts.
 Our algorithm is non-backtracking without extra efforts.
 On the other hand, (Pager1977) has to define ``compatibility'' carefully. \\
 (5) There is no guarantee that (Pager1977) will find the minimal LR(1) machines.
 (6) His advantage is that his algorithm does not have to construct the big LR(1) machines.
 In contrast our algorithm is easier to understand.
 As the memory constraint is less and less severe on modern computers,
 the advantage of his algorithm is less and less valuable. \\
 (7) The algorithm in (Pager1977) still merges {\it similar states} as new states are generated.
 (8) (Pager1977) tried hard to reduce the size of the parse table. However, since minimizing LR(1) machines
 is NP-hard, the algorithms are considered ``heuristics''.
 There is no clear evidence which algorithms will produces the smaller LR(1) machines.

 (Henhir) is an implementation of (Pager1977) algorithm.

 (21 Pager Lane tracing) Pager, D.: The lane tracing algorithm for constructing LR(k)
 parsers and ways of enhancing its efficiency. Information Sciences (to appear)  \\
 (21 Pager Lane tracing)The latter (21) employs an LR(O) algorithm initially, and then, if the
 grammar is non-LALR, splits states so as to remove conflicts.

\ \



 (Tribble2006) D.R. Tribble, the Honalee LR(k) Algorithm, Revision 1.0, 2006-04-27. \\
 (Tribble2006) is not mature. The key points in the paper is unclear.
 The presented algorithm and examples are about LR(1).
 Almost no discussion of LR(k).
 Nothing is said why it produces a ``minimal'' machine.
 Nothing is said about complexity.


 Weakness of our method: need to create the whole LR(1) machine first, which is a big machine, and then merge similar states.

\ \

It is known that every language that admits an LR(k) parser also admits an LALR(1) parser [9] but at the expense of exponential explosion of states. This is related to our NP-hardness conclusion.


}

\end{document}
